\documentclass[conference]{IEEEtran}
\usepackage[hyphens]{url}
\usepackage{hyperref}
\hypersetup{breaklinks=true,colorlinks,allcolors=blue}
\usepackage{doi}

\IEEEoverridecommandlockouts
\usepackage{cite}
\usepackage{caption}
\usepackage{algorithmic}
\usepackage{graphicx}
\usepackage{textcomp}
\usepackage{xcolor}
\usepackage{balance}
\usepackage[left=1.62cm,right=1.62cm,top=1.9cm]{geometry}

\def\BibTeX{{\rm B\kern-.05em{\sc i\kern-.025em b}\kern-.08em
    T\kern-.1667em\lower.7ex\hbox{E}\kern-.125emX}}
\begin{document}

\newcommand{\adam}[1]{{\color{magenta}Adam: #1}}
\newcommand{\mamun}[1]{{\color{red}Mamun: #1}}

\title{Introducing Axlerod: An LLM-based Chatbot for Assisting Independent Insurance Agents}

\author{\IEEEauthorblockN{Adam Bradley, John Hastings and Khandaker Mamun Ahmed$^*$}
 \IEEEauthorblockA{\textit{The Beacom College of Computer and Cyber Sciences, Dakota State University} \\
Madison, SD, USA \\ $^*$Corresponding Author: khandakermamun.ahmed@dsu.edu}
}

\maketitle

\begin{abstract}
The insurance industry is undergoing a paradigm shift through the adoption of artificial intelligence (AI) technologies, particularly in the realm of intelligent conversational agents. Chatbots have evolved into sophisticated AI-driven systems capable of automating complex workflows, including policy recommendation and claims triage, while simultaneously enabling dynamic, context-aware user engagement. This paper presents the design, implementation, and empirical evaluation of Axlerod, an AI-powered conversational interface designed to improve the operational efficiency of independent insurance agents. Leveraging natural language processing (NLP), retrieval-augmented generation (RAG), and domain-specific knowledge integration, Axlerod demonstrates robust capabilities in parsing user intent, accessing structured policy databases, and delivering real-time, contextually relevant responses. Experimental results underscore Axlerod’s effectiveness, achieving an overall accuracy of $93.18$\% in policy retrieval tasks while reducing the average search time by $2.42$ seconds. This work contributes to the growing body of research on enterprise-grade AI applications in insurtech, with a particular focus on agent-assistive rather than consumer-facing architectures.
\end{abstract}

\begin{IEEEkeywords}
Large language models, Generative
AI, Prompt engineering, Chatbots, AI for insurance agents, Insurtech
\end{IEEEkeywords}

\section{Introduction}

The insurance industry plays a vital role in both the financial stability and economic development of society~\cite{arena2008does}. It contributes by mitigating risk, providing financial security and facilitating economic activity~\cite{arrow1996theory}. There are more than $4,100$ property and casualty insurance carriers in the United States with a combined revenue of more than USD $1$ trillion \cite{ac_editor_naic_2025, ibisworld_inc_property_nodate}.  The US insurtech sector, made up of firms that provide technology services to this industry, comprises approximately $1,500$ companies with revenues exceeding \$$9.2$ billion in 2024 \cite{berenguer_revealed_2024, dimension_market_research_insurtech_2024}. 
In order to improve efficiency, the insurance industry has a long-standing tradition of leveraging technological innovation to improve operations~\cite{eling2018impact}, improve decision making~\cite{frees2014predictive}, and enrich customer experiences~\cite{eckert2022managing}. Over the past few decades, this evolution has progressed from rule-based automation systems~\cite{meyer1990strategic} to the integration of predictive analytics~\cite{de2008generalized} and more recently, advanced AI methodologies such as machine learning (ML)~\cite{blier2021machine} and generative AI~\cite{balona2024actuarygpt}. A particularly transformative development in this trajectory is the emergence of large language models (LLMs), which have shown remarkable potential to automate and augment insurance-related tasks including underwriting~\cite{munichre}, claims adjudication~\cite{swissre}, fraud detection~\cite{islayem2025using}, and policyholder engagement~\cite{patil2024chat}.

Although much of the current discourse surrounding AI-powered chatbots in insurance has focused on their utility in customer-facing applications, this paper discusses the deployment of LLM-based chatbots as tools to support insurance agents rather than customers. This shift is not only pragmatic but strategically advantageous. Insurance producers, by virtue of their domain expertise, are better positioned than consumers to interpret nuanced policy language, validate output accuracy and to detect hallucinated responses which are the common failure modes of generative AI systems. Therefore, using chatbots in agent-facing roles offers a safer yet highly impactful application of LLMs within regulated service environments. Despite this compelling rationale, limited work conducted on chatbot deployment in the insurance sector reveals a significant gap where agent-assistive AI systems received minimal attention. Therefore, this research gap provides an opportunity for innovation and inquiry, particularly in enhancing agents’ ability to retrieve policy information, cross-reference documents, and provide accurate and prompt client support. Given that agents serve as both the point of sale and a crucial channel for customer support, enhancing their efficiency would yield benefits for insurance customers and overall business operations. We believe that the implementation of an AI assistant could improve agent productivity and thus carrier profitability by efficiently locating the necessary information for the agent and thereby reducing the time overhead.

In this study, we collaborated with Safety Insurance, a regional insurance carrier specializing in personal and commercial property and casualty insurance across Massachusetts, Maine, and New Hampshire \cite{safety_insurance_group_safety_2025}. Over the past two years, Safety Insurance has been actively investigating the potential applications of LLMs as part of a broader AI adoption strategy. Our partnership with Safety was established to build a practical prototype that could augment their agent operations and evaluate the feasibility and reliability of LLM-driven assistants in real-world insurance workflows.  Therefore, the primary objective of this research is to develop a chatbot prototype capable of assisting Safety Insurance agents with policy and coverage-related inquiries. We have named this chatbot ``Axlerod" which is specifically designed to help agents review client coverage, query large-scale policy databases, retrieve relevant documentation and respond to complex customer questions with contextual precision. Our approach leverages retrieval-augmented generation (RAG), intent recognition, and structured tool integration to optimize performance and reliability. Finally, our objective is to assess whether current LLM technology can contribute meaningfully to improving operational efficiency, reducing response times, and improving customer service standards in the insurance context. While the increasing use of AI across industries raises strong and legitimate concerns about job displacement, we intend Axlerod explicitly as a support tool designed to empower rather than replace insurance agents. By automating repetitive information search and retrieval tasks and minimizing the risk of human error in data lookup, Axlerod allows agents to focus on relationship building, client consultation and decision making functions that continue to require human judgment and empathy. In this way, the chatbot serves as an augmentation layer rather than a replacement, keeping humans in the loop and reinforcing the central role of human expertise in the value proposition of the insurance industry as a whole and independent insurance agents in particular.

The contributions of our work are summarized as follows:

\begin{itemize}
    \item   This study is the first to explore the utility of LLM-based conversational agents as support tools for insurance agents and enhance their interactions capability with customers.
    \item We describe a fully operational end-to-end system capable of querying a multi-source insurance data stack, encompassing hundreds of thousands of live policies and associated documents.
    \item We contribute empirical evidence quantifying the accuracy and robustness of LLM tool use in structured insurance tasks and thereby informing future applications and benchmarks in the domain.
\end{itemize}

The remainder of this paper is structured as follows. Section \ref{related_works} provides a comprehensive review of existing research and prior developments related to chatbot applications, with particular emphasis on their role within the insurance domain. Section \ref{System Design and Implementation} describes the system architecture in detail, including the design principles, implementation choices, and the development process of the proposed AI-driven chatbot agent. Section \ref{Results and Discussion} presents the experimental results. Section \ref{future work} outlines potential avenues for future research, highlighting directions that can extend the scope and impact of this study. Finally, Section \ref{Conclusions} summarizes the key contributions of this work.

\section{Related Work} \label{related_works}
In recent years, with the advancements of AI technologies, especially LLMs, the insurance industry has been leveraging AI to analyze real-time data.  These technologies provide more precise and personalized risk assessments compared to traditional demographic-based methods, thereby making Usage-Based Insurance (UBI) viable \cite{holland2021huk}. As a result, major insurance software providers such as Guidewire are also integrating LLM systems into their core operations \cite{SateeshReddy2024-SATACP-2}. Moreover, early chatbot implementations in insurance, such as IntelliBot, have also demonstrated timely and relevant responses to customer inquiries, thus improving customer satisfaction and operational efficiency \cite{nuruzzaman2020intellibot}. Despite these advancements, AI adoption in insurance faces critical technical and operational challenges. These include ensuring accurate and unbiased data collection, managing privacy concerns, and providing transparency in AI-driven decision-making processes to maintain regulatory compliance and consumer trust \cite{holland2021huk}. As trust is crucial for effective AI deployment, addressing AI-induced risks such as hallucinations, biases, and goal misalignment is significant. The authors in \cite{weerawardhana2024exploring} mention that uncertainty and misaligned objectives between AI systems and users reduce trust which highlights the necessity for careful alignment of chatbot functionalities with user expectations.  

The existing body of research predominantly emphasizes the application of chatbots in customer-facing roles, leaving the potential for agent-facing chatbots unexplored. Human insurance agents continue to perform critical functions within sales, client management, and retention, areas where automated systems have limited effectiveness. The authors in \cite{singh2022disruptive} argue that equipping human agents with improved informational tools significantly enhances their productivity and client retention metrics. In light of this, our research shifts the focus towards developing AI chatbots designed explicitly to support insurance agents. Such agent-assistive systems necessitate distinct functional requirements, including reliable retrieval of policy information, dynamic document generation, efficient management of coverage changes, and streamlined access to internal knowledge bases \cite{chudleigh_complete_2024}. Ensuring these features are correctly implemented facilitates prompt and accurate client interactions, addressing an underexplored but essential area of research.

Beyond functionality, ethical considerations significantly influence the design and deployment of AI-based chatbots within insurance contexts. Central to ethical deployment are principles of transparency ensuring users clearly understand when they are interacting with automated systems and accountability, mandating human oversight mechanisms to handle errors or adverse outcomes generated by AI systems \cite{sheir2025adaptable}. The authors in \cite{crowder2024ai} propose a structured ethical decision-making framework for AI deployment that includes careful observation, alternatives evaluation, decision making, action, and post-action reflection, which further emphasizes the necessity of deliberate design and oversight. Additionally, the authors in \cite{gupta_ethical_2022} highlight the importance of continuous monitoring and refinement of chatbot systems to address biases, ensure fairness, and adapt to changing circumstances. The regulatory landscape is also evolving to address the implications of AI-driven insurance technologies. In 2020, the National Association of Insurance Commissioners (NAIC) established foundational guidelines in its Principles on Artificial Intelligence which emphasize fairness, ethics, transparency, accountability, compliance and security in AI deployment \cite{national_association_of_insurance_commissioners_national_2020}. Concurrently, evolving legislative frameworks across the United States further underline the necessity for insurance technology providers to comply with emerging regulatory standards, creating an additional imperative for careful system design and rigorous compliance monitoring \cite{gosztonyi2024online}.

Implementing AI chatbots in agent-supporting roles demands thoughtful and strategic engineering practices. The authors in \cite{erik_schluntz_building_2024} emphasize simplicity and modularity in the design of AI-driven systems, distinguishing between predefined workflow automation and flexible, agentic AI systems capable of dynamically adapting to various tasks. They suggest using simple, composable, and lightweight patterns rather than complex frameworks when developing AI agents and provide a list of guidelines for developing of AI agents, several of which we have adopted in Axlerod's development. An alternative chatbot design approach, Cache-Augmented Generation (CAG), proposed in \cite{chan2025don}, feeds LLMs comprehensive datasets at the outset, enabling response generation without additional Application Programming Interface (API) calls. Although this approach offers efficiency advantages for document-based queries, its effectiveness diminishes when applied to highly dynamic and complex databases such as those used in policy management and client services. In contrast to this study, our system accesses the dynamic and complex databases for policy management and client services. A critical implementation consideration involves effective tool orchestration by chatbots. The authors in \cite{gao2024efficient} proposed a chain-of-abstraction reasoning method, where LLMs are trained to use abstract reasoning to effectively manage sequential tool usage for complex tasks. Similarly, the authors in \cite{wu2024avatar} developed AvaTaR, an automated training framework for AI agents that enables the generalizable and efficient tool usage across diverse task contexts. These methodologies underscore the importance of training AI models not only in response generation but also in structured reasoning for reliable tool integration capabilities essential in insurance workflows, where precision and accuracy are paramount.

The existing literature highlights the current state of AI integration in insurance, but leaves significant gaps. particularly in agent-supporting chatbot applications, and underscores the technical and practical considerations essential for successful system implementation. Our research addresses these identified gaps, contributing novel insights into the feasibility, design, and practical integration of agent-assistive chatbot technologies within insurance operations.

\section{System Design and Implementation} \label{System Design and Implementation}

\subsection{Agent Design} \label{agent_design}

\begin{figure}[!ht]
    \centering
    \includegraphics[width=0.9\linewidth]{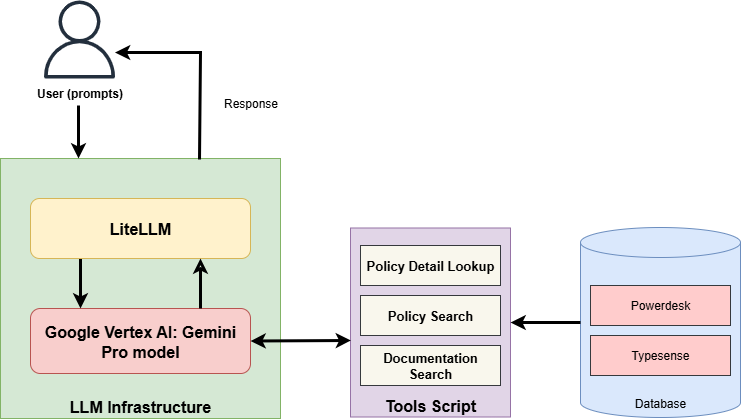}
    \caption{Overall architecture of Axlerod. The system connects an LLM (Google Gemini 2.5 Pro) with Safety Insurance’s internal data sources through a lightweight middleware layer. Three primary tools: policy detail, policy search, and documentation search are exposed to the LLM for structured retrieval.}
    \label{fig1}
\end{figure} 

At its core, Axlerod is a lightweight application that provides the interface between the LLM hosted on Google's cloud infrastructure and Safety Insurance’s internal systems and databases. Access to these data sources is provided by three functions (listed below), made available to the LLM as ``tools" in LLM jargon. Inputs from the agent are forwarded by this application to the LLM along with structured instructions and metadata detailing the available tools. Based on these inputs, the LLM either directly generates an appropriate response or issues a structured ``tool call", an instruction for the core application to execute a specific function to retrieve policy information. Fig.~\ref{fig1} illustrates this architecture, depicting the data flow from user input (prompts) to the LLM, available tool calls, and generation of final responses. Currently, Axlerod has access to these tools:
\begin{enumerate}
    \item policy\_detail: Retrieves comprehensive details for a given policy number.
    \item policy\_search: Queries a Typesense search engine to find a policy given a policyholder's name or address.
    \item documentation\_search: Allows the chatbot to search and access policy documentation.
\end{enumerate}

Upon receiving a tool call, the system performs the corresponding function, retrieves data, and returns the results to the LLM. The LLM subsequently generates a refined response using the newly acquired data, which is then forwarded to the user interface. The chatbot interface is integrated within the agent-facing web application platform, enabling context-aware interactions. For example, when an agent reviews a particular policy, Axlerod is informed about this context, allowing it to accurately interpret and respond to agent queries referring to ``this policy."

In this system, we utilize Google's Gemini 2.5 Pro~\cite{comanici2025gemini}, a state-of-the-art generative model hosted on Google's Vertex AI infrastructure. Initially, the system employed the 80 billion-parameter Llama 3.1 model \cite{schmid2024llama}, as it was open-weight and could be run on internal hardware; however, the transition to Gemini was motivated by its advanced capabilities and alignment with the company's existing adoption of Google Cloud services. Integration between Gemini and the backend application occurs via LiteLLM~\cite{litellm}, a proxy service that translates API requests from OpenAI's standard format into compatible calls for Google's API, while managing authentication with Vertex AI. To minimize complexity, our system's interaction between the LLM and toolsets was initially developed without reliance on third-party frameworks. In order to avoid the complications among different functions, we designed an independent ``microframework" named Smoltalk.

\subsection{Data Description}

At this stage of development, we specifically focused on granting Axlerod access to three distinct categories of data: 

\begin{itemize}
    \item The policy database includes a comprehensive inventory of about 730,000 personal and commercial policies. The features of this database include policy number, policy types, policyholder name and address, policy effective and termination dates, and billing dates, installment plans, and upcoming payment amounts.
    \item Information on coverages related to those policies, including limits and vehicles covered; and claims, including amounts, descriptions, and related documentation.
    \item The documentation database, comprises details of company policies and procedures which are largely composed of unstructured text, and totaling about 400 MB of data. These documents are specific to the state and type of the policy in question. They include both company-specific manuals and standard insurance industry documentation.
\end{itemize}

\subsection{Tools and Technologies}

We use several open-source tools in the development of Axlerod. Typesense~\cite{typesense}, a search engine serves as the database supporting policy search. Python 3.11 functions as the foundational programming language and is augmented by third-party libraries such as httpx to enable interaction with web-based APIs. The FastAPI web application framework~\cite{fastapi} delivers the Web API to the chat interface while Smoltalk, a microframework that we developed for agentic LLM handles interactions with the LLM. Smoltalk simplifies interactions between LLMs and external tools by offering an OpenAI-compatible API that embeds easily in FastAPI applications. Developers define tool functions as Python class methods which the Smoltalk Toolbox converts into metadata for seamless LLM integration. Open WebUI~\cite{openwebui} plays an integral role in the early stages of development due to its extensibility and robust feature set, including an API compatible with OpenAI’s. LiteLLM, an LLM gateway server, enables seamless access to multiple LLMs from various providers through a unified API. Finally, the Tampermonkey~\cite{tampermonkey} userscript manager is used to prototype the embedded chat interface.

\section{Experimental Analysis} \label{Results and Discussion}

We evaluated our LLM-powered chatbot in a controlled, internal environment using a set of scripted and ad-hoc queries representative of agent workflows. Moreover, to test the developed framework, we include both manual testing using the integrated chat widget and automated tests using Python's unittest module to send questions directly to Axlerod's API and compare the results with known truth. The test policies are chosen at random from the current active policy database.

\subsection{Accuracy}
To measure the accuracy of the chatbot, we provide a number of specific queries regarding the policies. The queries include: (1) What is the due date and amount due for (policy number)? (2) What is (policyholder)'s auto policy number? (3) What vehicles are covered by (policy number)? (4) What bill plan is this policy on?

\begin{figure}[h]
    \centering
    \includegraphics[width=0.9\linewidth]{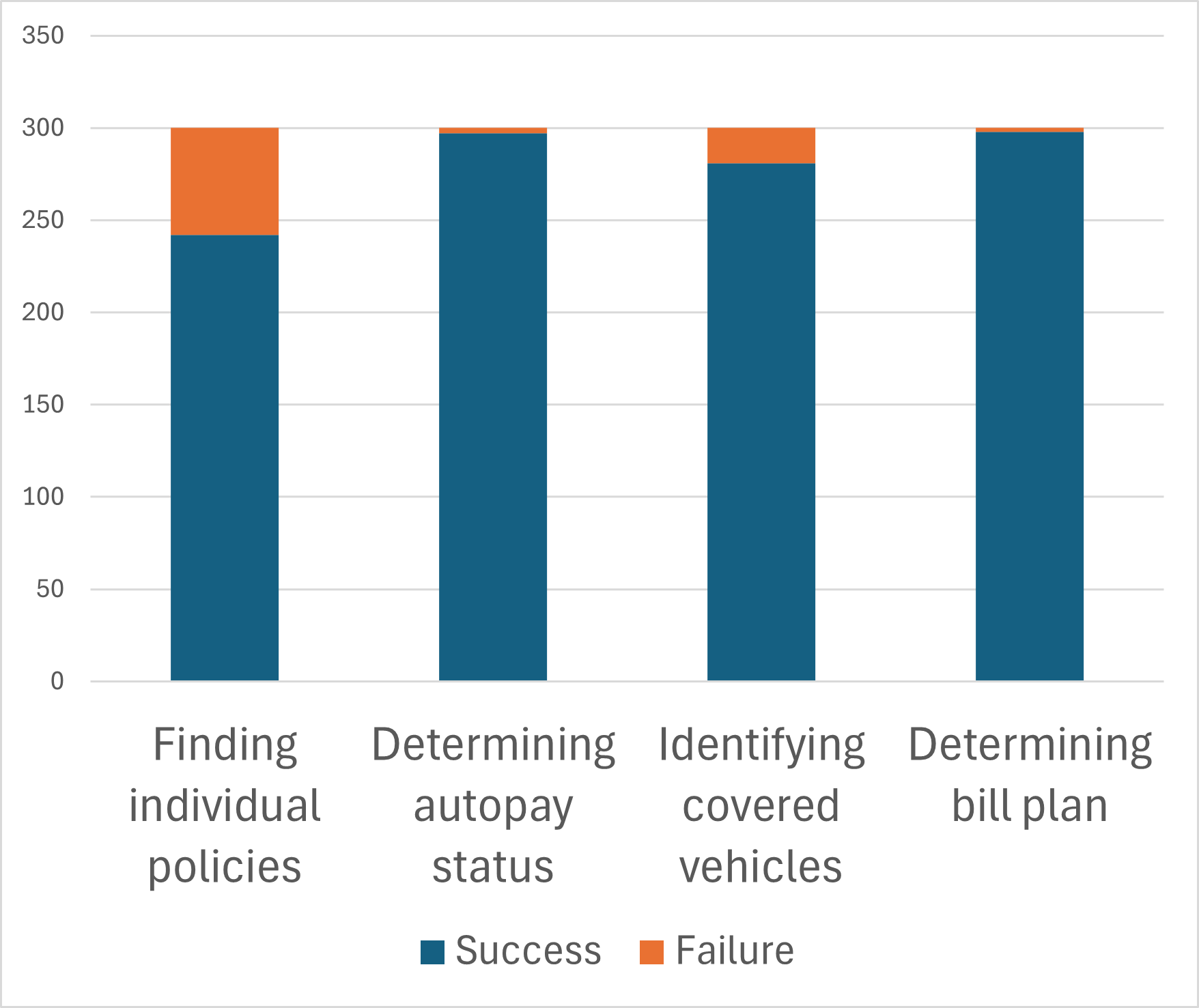}
    \caption{The accuracy of query responses by the developed AI chatbot.}
    \label{fig2}
\end{figure}

Moreover, we use multiple variations of each of these questions, representing different language styles and registers, to test the model's ability to understand a wider variety of language registers and phrasings. The system achieves an accuracy of 80.7\% in correctly identifying policy numbers. The majority of these errors are attributed to ambiguities arising from common or frequently occurring names. By design, Axlerod prompts the user for additional input when an initial query yields more than five results to refine and narrow the scope of the search. The accuracy when determining whether a policy is enrolled in the AutoPay program is 99.0\%, and to identify covered vehicles using an automobile policy number is 93.7\%. Moreover, the accuracy when asked for a policy's bill plan was 99.3\%. Axlerod achieved an overall accuracy of 93.18\%, correctly answering 1,118 out of 1,200 queries across four task types. In Fig. \ref{fig2}, we show the accuracy results where the highest accuracy is achieved in determining bill plan queries followed by finding individual policies.

\begin{table}[!t]
\centering
\caption{Performance comparison of human agent with and without chatbot assistance. Each of two agents completed four different tasks with eight different policy numbers in this scenario. Average time to locate the requested information in each task is given in seconds.}
\label{tab:table1avg}   
\begin{tabular}{l | p{65pt} | p{60pt}}  
\hline
\textbf{Tasks} & \textbf{Search Time Without Axlerod (s)} &\textbf{ Search Time With Axlerod (s)} \\
 \hline
Find Policy Number & 7.6 & 5.3 \\
Get Autopay Eligibility & 5.4& 5.2\\
Get Covered Vehicles & 14.0 & 5.1 \\
Get Bill Plan & 5.3& 4.9 \\ \hline
\textbf{Average Time} & 7.55 & 5.13 \\

\end{tabular}
\end{table}

\subsection{Evaluation of Time Savings}
To evaluate time savings, we assign two users to complete four distinct tasks using two policies each, measuring the time required to perform each task both with and without Axlerod. Axlerod retrieves covered vehicle information significantly faster than a user relying solely on the default search interface. For other query types, Axlerod demonstrates minimal or no time advantage when compared to experienced users. However, when compared with users unfamiliar with the application's search functionality and page layout, Axlerod typically retrieves information several seconds faster. Table \ref{tab:table1avg} presents the time (in seconds) required to complete each task. With simple tasks, where the user already has access to the policy number, Axlerod only marginally improves search times. With the help of Axlerod, human agents perform their task significantly faster comparing with the traditional way. On average, without the chatbot a human agent takes $7.55$ seconds where with chatbot the agent takes $5.13$ seconds, and the $2.42$ seconds faster search time would play a significant role in serving a large number of customers. With more complex tasks, where a user would need to navigate multiple screens or search for a user by name, Axlerod is notably faster. 

\subsection{Cost}
Based on the costs reported by LiteLLM during the testing, we calculate that an average answer from Axlerod costs approximately \$0.0075. However, the effective cost is significantly lower considering the efficiency improvement of the agents by using Axlerod for serving customers. 

\subsection{Example Conversation}
In this section, we present an illustrative example of an interaction between the agent and Axlerod. The following transcript demonstrates how the system accurately retrieves and interprets policy-related information through iterative clarification. Specifically, the chatbot first requests additional identifying details, subsequently isolates the correct policy record, and finally identifies the associated billing plan. All personally identifying information has been removed to preserve confidentiality.

User: What is (redacted)'s auto policy number?

Axlerod: I found several auto policies for (redacted). Could you provide more information to help me narrow down the search, such as a city or address?

User: Sure, I'm looking for the (redacted) in Fall River.

Axlerod: I found one auto policy for (redacted) in Fall River: (Redacted: correct policy number, named insureds, address, and powerdesk link.)

User: What bill plan is that policy on?

Axlerod: This policy is on a 12-Pay bill plan.

\section{Future Work} \label{future work}
Future work will focus on improving accuracy, reducing the impact of fragmented and difficult-to-access internal data sources, and refining the user interface for seamless integration within PowerDesk. Expanded testing with practicing insurance agents will be critical for validating real-world utility, assessing agent trust, and measuring efficiency in both human-in-the-loop and fully automated workflows. To strengthen methodological rigor, we also plan richer empirical validation through broader user studies and more complex multi-step tasks. In addition, we will further position our lightweight Smoltalk framework against established alternatives such as LangChain and Microsoft’s Semantic Kernel, clarifying its advantages in simplicity and ease of integration.

\section{Conclusion} \label{Conclusions}
This study highlights the transformative role of artificial intelligence in the insurance industry, demonstrating how conversational agents such as Axlerod can enhance operational efficiency by assisting independent insurance agents with policy retrieval and workflow automation. Through the integration of natural language processing, retrieval-augmented generation, and domain-specific knowledge, Axlerod shows promise as a robust and context-aware assistant. The findings underscore both the opportunities and challenges associated with deploying AI-driven systems in high-stakes domains.

\section*{Acknowledgments}
The authors express their sincere appreciation to colleagues at Safety Insurance, particularly Keith Carangelo, Vasili Eliopoulos, and Josh Silver, for providing technical assistance and facilitating access to essential policy data. Additionally, the authors acknowledge the use of generative AI tools exclusively to enhance grammatical accuracy and manuscript readability. No original content or findings were generated by these AI tools beyond editorial refinements.

\section*{Availability}

Additional notes regarding testing, and Axlerod's system prompt, are available on Github at \url{https://github.com/atbradley/axlerod-agent-assistant-paper}. Smoltalk is available at \url{https://github.com/atbradley/smoltalk}.

\section*{Conflicts of interest}
The authors declare no conflict of interest.

\balance
\bibliographystyle{IEEEtranDOIandURLwithDate}
\bibliography{new_ref}

\end{document}